%% file: arxiv.tex
\documentclass[runningheads]{llncs}

\usepackage{style}
\usepackage{abbrv}

\usepackage{graphicx}
\usepackage{booktabs}
\usepackage[accsupp]{axessibility}  
\usepackage{tikz}
\usepackage{color}
\usepackage{nicefrac}
\usepackage{adjustbox}
\usepackage[accsupp]{axessibility}  
\usepackage{pgfplots}
\pgfplotsset{compat=1.7}
\usepackage{multirow}

\newcommand{\mj}{$\mathcal{J}$}
\newcommand{\mf}{$\mathcal{F}$}
\newcommand{\mjf}{$\mathcal{J}\&\mathcal{F}$}

\input{mlVecMat}

\definecolor{LightCyan}{rgb}{0.88,1,1}
\usepackage{hyperref}

\begin{document}

\title{Efficient Video Object Segmentation via Modulated Cross-Attention Memory} 
\author{Abdelrahman Shaker\inst{1}\and
Syed Talal Wasim\inst{2,3} \and
Martin Danelljan\inst{4} \and
Salman Khan\inst{1} \and
Ming-Hsuan Yang\inst{5,6} \and
Fahad Shahbaz Khan\inst{1,7}
}

\authorrunning{Shaker et al.}

\institute{Mohamed Bin Zayed University of AI \and
University of Bonn \and
Lamarr Institute for ML and AI \and
ETH Zürich \and
University of California, Merced \and
Google Research \and
Linkoping University}

\maketitle

\begin{abstract}
  Recently, transformer-based approaches have shown promising results for semi-supervised video object segmentation. However, these approaches typically struggle on long videos due to increased GPU memory demands, as they frequently expand the memory bank every few frames. We propose a transformer-based approach, named MAVOS, that introduces an optimized and dynamic long-term modulated cross-attention (MCA) memory to model temporal smoothness without requiring frequent memory expansion. The proposed MCA effectively encodes both local and global features at various levels of granularity while efficiently maintaining consistent speed regardless of the video length. Extensive experiments on multiple benchmarks, LVOS, Long-Time Video, and DAVIS 2017, demonstrate the effectiveness of our proposed contributions leading to real-time inference and markedly reduced memory demands without any degradation in segmentation accuracy on long videos. Compared to the best existing transformer-based approach, our MAVOS increases the speed by 7.6$\times$, while significantly reducing the GPU memory by 87\% with comparable segmentation performance on short and long video datasets. Notably on the LVOS dataset, our MAVOS achieves a $\mathcal{J \& F}$ score of 63.3\% while operating at 37 frames per second (FPS) on a single V100 GPU. Our code and models will be publicly available at: \url{https://github.com/Amshaker/MAVOS}.
\end{abstract}

\section{Introduction}
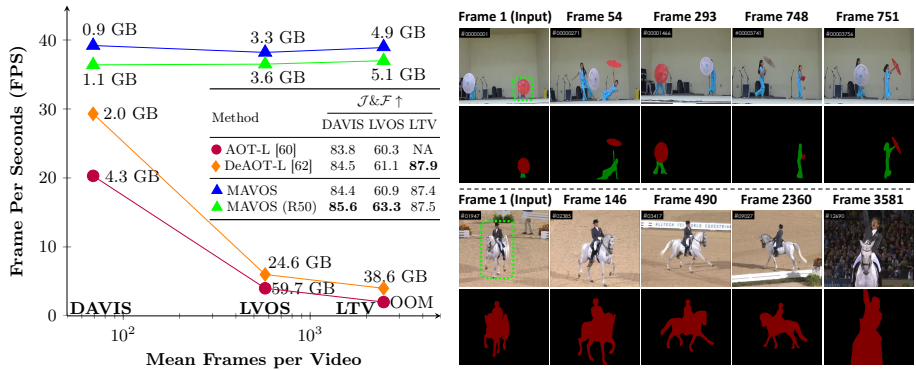
\begin{figure}[t!]
  \centering
  \begin{subfigure}[b]{0.5\textwidth}
    \centering
    
\resizebox{\textwidth}{!}{
	\begin{tikzpicture}
\begin{semilogxaxis}[
axis lines = left,
ymin =0, ymax= 45,
xmin =50, xmax = 5000,
xlabel= \textbf{Mean Frames per Video},
ylabel= \textbf{Frame Per Seconds (FPS)},
]

\coordinate (legend) at (axis description cs:0.95,0.55);

\addplot[only marks, mark=otimes*, purple, mark size=3pt]
    coordinates {
    (69, 20.3)
    (574, 4)
    (2470, 2)
    };\label{plot:AOT}

\node[right] at (axis cs: 74, 20.3) { 4.3 GB};
\node[right] at (axis cs: 574, 4) { 59.7 GB};
\node[right] at (axis cs: 2470, 2) { OOM};

\addplot[sharp plot, purple]
    coordinates {
    (69, 20.3)
    (574, 4)
    (2470, 2)
    };

\addplot[only marks, mark=diamond*, orange, mark size=3.5pt]
    coordinates {
    (69, 29.3)
    (574, 6)
    (2470, 4)
    };\label{plot:DEAOTL}

\node[above] at (axis cs: 110, 27.7) {2.0 GB};
\node[above] at (axis cs: 884, 6) {24.6 GB};
\node[above] at (axis cs: 2870, 4) {38.6 GB};

\addplot[sharp plot, orange]
    coordinates {
    (69, 29.3)
    (574, 6)
    (2470, 4)
    };

\addplot[only marks, mark=triangle*, blue, mark size=3.9pt]
    coordinates {
    (69, 39.2)
    (574, 38.2)
    (2470, 39.0)
    };\label{plot:Ours}

\addplot[only marks, mark=triangle*, green, mark size=3.9pt]
    coordinates {
    (69, 36.4)
    (574, 36.5)
    (2470, 37.0)
    };\label{plot:Ours_R50}

\addplot[sharp plot, green]
    coordinates {
    (69, 36.4)
    (574, 36.5)
    (2470, 37.0)
    };
    
\node[above] at (axis cs: 85, 39.8) {0.9 GB};
\node[above] at (axis cs: 674, 38.7) {3.3 GB};
\node[above] at (axis cs: 2970, 39.4) {4.9 GB};

\node[above] at (axis cs: 85, 32.5) {1.1 GB};
\node[above] at (axis cs: 674, 32.8) {3.6 GB};
\node[above] at (axis cs: 2970, 33.3) {5.1 GB};

\addplot[sharp plot, blue]
    coordinates {
    (69, 39.2)
    (574, 38.2)
    (2470, 38.9)
    };
\end{semilogxaxis}

\begin{semilogxaxis}[
    axis lines = none,
    xmin = 50, xmax = 5000,
    ymin = 0, ymax = 1,
    axis on top,
]
\node[align=center] at (axis cs: 76, 0.025) {\textbf{DAVIS}};
\node[align=center] at (axis cs: 574, 0.025) {\textbf{LVOS}};
\node[align=center] at (axis cs: 1760, 0.025) {\textbf{LTV}};
\end{semilogxaxis}
\node[draw=none,fill=none, anchor= north east, xshift=15pt, yshift=33pt] at 
(legend){\adjustbox{width=0.72\textwidth}{

\begin{tabular}{lccc}
        \toprule
        \multirow{3}{*}{Method} & \multicolumn{3}{c}{$\mathcal{J \& F}\uparrow$ } \\
        \cmidrule(lr){2-4}
        & DAVIS  & LVOS & LTV\\
        \midrule
        \ref{plot:AOT} AOT-L~\cite{yang2021aot} & 83.8 & 60.3 & NA \\
         \ref{plot:DEAOTL} DeAOT-L~\cite{yang2022deaot} & 84.5 & 61.1 & \textbf{87.9} \\
        \midrule
        \ref{plot:Ours} MAVOS & 84.4 & 60.9 & 87.4 \\
        \ref{plot:Ours_R50} MAVOS (R50)& \textbf{85.6} & \textbf{63.3} & 87.5 \\
        \bottomrule
    \end{tabular}
}};

\end{tikzpicture}
}
  \end{subfigure}%
  \begin{subfigure}[t]{0.5\textwidth}
    \centering
    \resizebox{\textwidth}{!}{\includegraphics[width=0.5\linewidth]{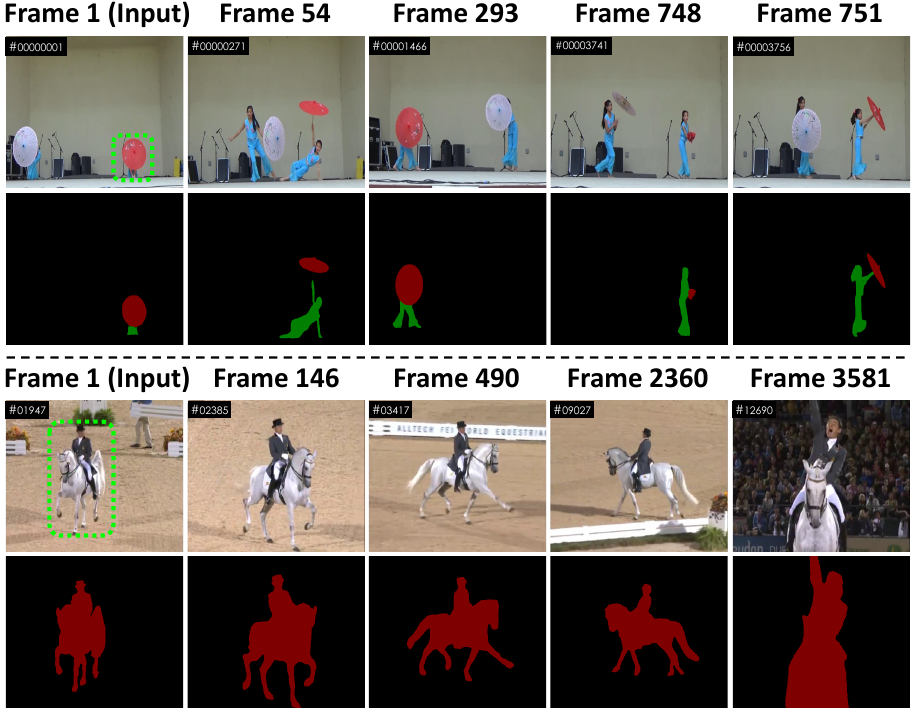}} 
  \end{subfigure}
  \caption{\textbf{Left:} Comparison of our proposed MAVOS with recent transformer-based methods using the same backbone in terms of speed (FPS) and mean frames per video, along with GPU memory consumption (in GB) and VOS performance on the graph. Recent transformer-based approaches exhibit a substantial reduction in speed and memory explosion for long videos, while MAVOS models maintain consistent speed without GPU memory issues and no significant performance degradation in both short (DAVIS~\cite{Pont-Tuset_arXiv_2017}) and long-video datasets (LVOS~\cite{voigtlaender2019feelvos}, LTV~\cite{liang2020afb-ubr}). FPS is measured using a V100 GPU. \textbf{Right:} MAVOS results on long videos from LVOS (top) and LTV (bottom) datasets, showcasing robust performance with more than 120 seconds for LVOS and more than 3500 frames for LTV. Additional results are presented in suppl. material.} 
  

  \label{fig:intro}
\end{figure}

Video object segmentation (VOS) is a challenging problem that received significant attention in recent years and has numerous real-world applications, including autonomous driving~\cite{janai2021computer}, video editing~\cite{oh2018fastRGMP}, and augmented reality~\cite{ngan2011video}. The problem involves tracking specific objects in an image sequence to gain a deeper understanding of how these objects interact with each other. 
In the semi-supervised VOS setting, the goal is to accurately identify and track the specified targets throughout the entire video, given the reference frame containing one or more object masks. Due to the inherent nature of the VOS problem, it is desired to develop methods that are accurate and operate in real-time, while maintaining a small memory footprint, especially when processing long videos.

Most existing VOS methods can be roughly divided into template matching-based~\cite{yan2018osmn, chen2018pml, voigtlaender2019feelvos, yan2020cfbi, wang2019ranet}, recurrent-based~\cite{perazzi2017learningMaskTrack, hu2017maskrnn, ventura2019rvos,li2022recurrent}, memory-based~\cite{cheng2022xmem,oh2019videoSTM,seong2020kmn,cheng2021stcn}, and transformer-based~\cite{yang2021aot,yang2022deaot,duke2021sstvos,mei2021transvos}. Template matching-based approaches usually use predefined object templates to match and identify objects in the entire video. Such approaches may struggle with object deformation and variations in appearance~\cite{cheng2022xmem}. Further, they incur significant computation costs when applied to long videos, adding to their overall complexity. On the other hand, recurrent-based methods involve conveying information from the recent frames using hidden representations or masks for individual object instances at each step of the recurrence process. Although these methods can process single~\cite{xu2018youtube} as well multiple~\cite{ventura2019rvos} objects, they often encounter issues such as drifting, sensitivity to occlusions, and high computational demands, which can limit their suitability for practical real-time applications. 
Memory-based methods, exemplified by STM~\cite{oh2019videoSTM}, heavily rely on a memory network to save and retrieve target features from all past frames. Owing to the success of transformers in several vision domains, transformer-based approaches have been introduced for VOS. These approaches are usually based on applying an attention mechanism between current and past frames. Recent transformer-based VOS methods~\cite{yang2021aot, yang2022deaot} have explored hierarchical propagation, allowing for an association of multiple objects and improved VOS capabilities.

While the aforementioned transformer-based VOS approaches have shown promise by effectively encoding long-range dependencies between the target and past frames, these models are often designed for short-term videos and typically struggle with memory requirements when processing long-term videos (e.g., more than 60 seconds). This is due to the memory bank constantly increasing every few frames, which becomes prohibitive in case of long videos as the GPU memory can not meet the ever-increasing memory demands. 
For instance, Fig.~\ref{fig:intro} shows that the GPU memory requirements of the recent transformer-based VOS method~\cite{yang2022deaot}, DeAOT-L, increase dramatically from 2.0 GB to 24.6 GB, when increasing the number of frames from 69 to more than 500 per video. This results in reducing the speed by 6$\times$ (from around 30 to 5 FPS). A potential strategy to address this issue is to utilize a compact memory bank, as used in memory-based VOS methods~\cite{liang2020afb-ubr,cheng2022xmem}. However, we empirically observe such a strategy to lead to significant deterioration in accuracy on long videos likely due to struggling with low-level pixel matching. In this work, we tackle the challenge of the speed and GPU memory overflow in transformer-based methods for long videos. We explore an efficient design that maintains consistent memory and speed across various video lengths without sacrificing performance, especially on long videos.

\noindent \textbf{Contributions:}  
We propose an efficient transformer-based VOS approach, named MAVOS, that utilizes a novel optimized and dynamic long-term modulated cross-attention (MCA) memory. Our MCA effectively encodes the temporal smoothness from the past frames. It captures both local and global features, at various levels of granularity. Unlike existing transformer-based VOS designs, the proposed MCA avoids the need to expand memory and allows us to propagate target information based on the temporal changes of the past frames, without increasing memory usage. This substantially reduces the computational complexity, enabling real-time segmentation with consistent GPU memory cost, while preserving the characteristic accuracy of a transformer-based design. 

Experiments on three VOS benchmarks, LVOS~\cite{voigtlaender2019feelvos}, Long-Time Video (LTV)~\cite{liang2020afb-ubr}, and DAVIS 2017~\cite{Pont-Tuset_arXiv_2017}, reveal the merits of our proposed contributions for short and long videos. Our MAVOS significantly outperforms recent transformer-based VOS methods (see Fig.~\ref{fig:intro}) in terms of speed and GPU memory consumption, while achieving comparable accuracy. On the LTV dataset, MAVOS achieves a $\mathcal{J \& F}$ score of 87.4\% while operating at around 40 FPS and consuming 87\% less GPU memory compared to the best existing transformer-based VOS method~\cite{yang2022deaot}.

\section{Related Work}
\label{sec:related_work}
Video Object Segmentation (VOS) has received much attention over the past decade~\cite{VOS_survey}. Early VOS methods relied on traditional optimization methods and graph representations, such as~\cite{SeamSeg2014, VOSpredp2012, lu2020videoGraphMem}. Recent VOS methods~\cite{yang2021aot,yang2022deaot,cheng2022xmem,Hong_2023_ICCV} leverage deep neural networks for improved performance.

\noindent\textbf{Online Learning Approaches:} Online learning methods either train or fine-tune networks during test time~\cite{caelles2017oneOSVOS, voigtlaender2017onlineOnAVOS, maninis2018videoOSVOSS}. These involve fine-tuning the pre-trained segmentation networks during test time to guide the network's focus on the specified object. Examples include OSVOS~\cite{caelles2017oneOSVOS} and MoNet~\cite{xiao2018monet}, advocating the fine-tuning of pre-trained networks based on annotations from the first frame. Extending this strategy, OnAVOS~\cite{voigtlaender2017onlineOnAVOS} introduces an online adaptation mechanism. Similarly, MaskTrack~\cite{perazzi2017learningMaskTrack} and PReM~\cite{luiten2018premvos} further develop these techniques by integrating optical flow to aid in the propagation of the segmentation mask across consecutive frames. Despite recent improvements in efficiency~\cite{meinhardt2020make, robinson2020learningTargetModel, park2021learning, bhat2020learning}, these approaches still rely on online adaptation showing sensitivity to input and diminishing returns with increased training data. \\
\noindent\textbf{Template-based Approaches:} To avoid test-time fine-tuning, several works use the annotated frames as templates and explore techniques for alignment. OSMN~\cite{yan2018osmn} utilizes a network for extracting object embedding and another for predicting segmentation based on the derived embedding. 
PML~\cite{chen2018pml} focuses on learning pixel-wise embedding using the nearest neighbor classifier, whereas VideoMatch~\cite{hu2018videomatch} incorporates a matching layer to correlate pixels between the current and annotated frames in a learned embedding space. FEELVOS~\cite{voigtlaender2019feelvos} and CFBI(+)~\cite{yan2020cfbi, yang2021cfbip} enhance the pixel-level matching by including local matching with the previous frame. RPCM~\cite{xu2022rpcm} introduces a correction module to enhance the reliability of pixel-level matching. LWL~\cite{bhat2020learning} proposes employing an online few-shot learner to acquire the capability of decoding object segmentation. \\
\noindent\textbf{Memory-based Approaches:} Recent approaches ~\cite{liang2020afb-ubr,li2020fast,li2022recurrent} are based on pixel-level matching to propagate target information, with compact global context modules. Although these methods have constant memory requirements, they still struggle to keep track after a long period of target disappearance in challenging long-term videos. XMem~\cite{cheng2022xmem} introduces an architecture for long videos inspired by the Atkinson-Shiffrin memory model. It employs multiple interconnected feature memory stores, including a rapidly updated sensory memory, a high-resolution working memory, and a compact sustained long-term memory. Although XMem shows promising performance on the long-time video dataset~\cite {liang2020afb-ubr}, it struggles to generalize to other challenging long video datasets, such as LVOS~\cite{voigtlaender2019feelvos}. Recently, DDMemory~\cite{voigtlaender2019feelvos} has been proposed to effectively encode temporal context and maintain fixed-size memory using three complementary memory banks. \\
\noindent\textbf{Transformer-based Approaches:}
Methods that rely on attention mechanisms employ diverse strategies, such as similarity or template matching algorithms, to determine the memory frames~\cite{oh2019videoSTM, duarte2019capsulevos, zhang2020transductive,huang2020fastTemporalAggregation,ge2021video}. Numerous works have explored developing models to leverage local/pixel-to-pixel information, thereby enhancing mask quality. This improvement is achieved through different strategies, including the use of kernels~\cite{seong2020kmn}, optical flow~\cite{xie2021efficient, yu2022batman} and transformers~\cite{mao2021joint, lan2022learning, yu2022batman}. 
The recently introduced transformer-based method, AOT~\cite{yang2021aot}, tackles semi-supervised VOS in multi-object scenarios by introducing Associating Objects with Transformers (AOT). Unlike existing methods, AOT associates multiple targets into the same high-dimensional embedding space using an identification mechanism, enabling simultaneous processing of multiple object matching and segmentation decoding. A Long Short-Term Transformer is designed for hierarchical matching and propagation from the memory to the target. Although it achieves promising results on short video benchmarks, it struggles to achieve real-time speed. DeAOT~\cite{yang2022deaot} extends AOT and introduces decoupling features in hierarchical propagation, separating object-agnostic and object-specific embeddings in two branches based on the Gated Propagation Module (GPM). The design of GPM is based on efficient single-head attention, compared to the heavy multi-head attention of AOT. The recent DeAOT performs better than AOT since the problem of losing object-agnostic visual information is resolved using decoupling the propagation of the visual and ID embeddings using two independent branches. Although DeAOT achieves promising accuracy on multiple short video benchmarks~\cite{Pont-Tuset_arXiv_2017,xu2018youtube}, its long-term memory design is not suitable for long videos since it stores all past memory frames. This leads to ever-expanding memory and lower FPS when handling long-term videos with thousands of frames.

\section{Method}
\label{sec:method}
\noindent\textbf{Motivation}: 
To motivate our proposed method, we identify two key factors to consider when developing an efficient and accurate approach for VOS.

\begin{figure*}[t]
  \centering
    \includegraphics[width=.95\linewidth]{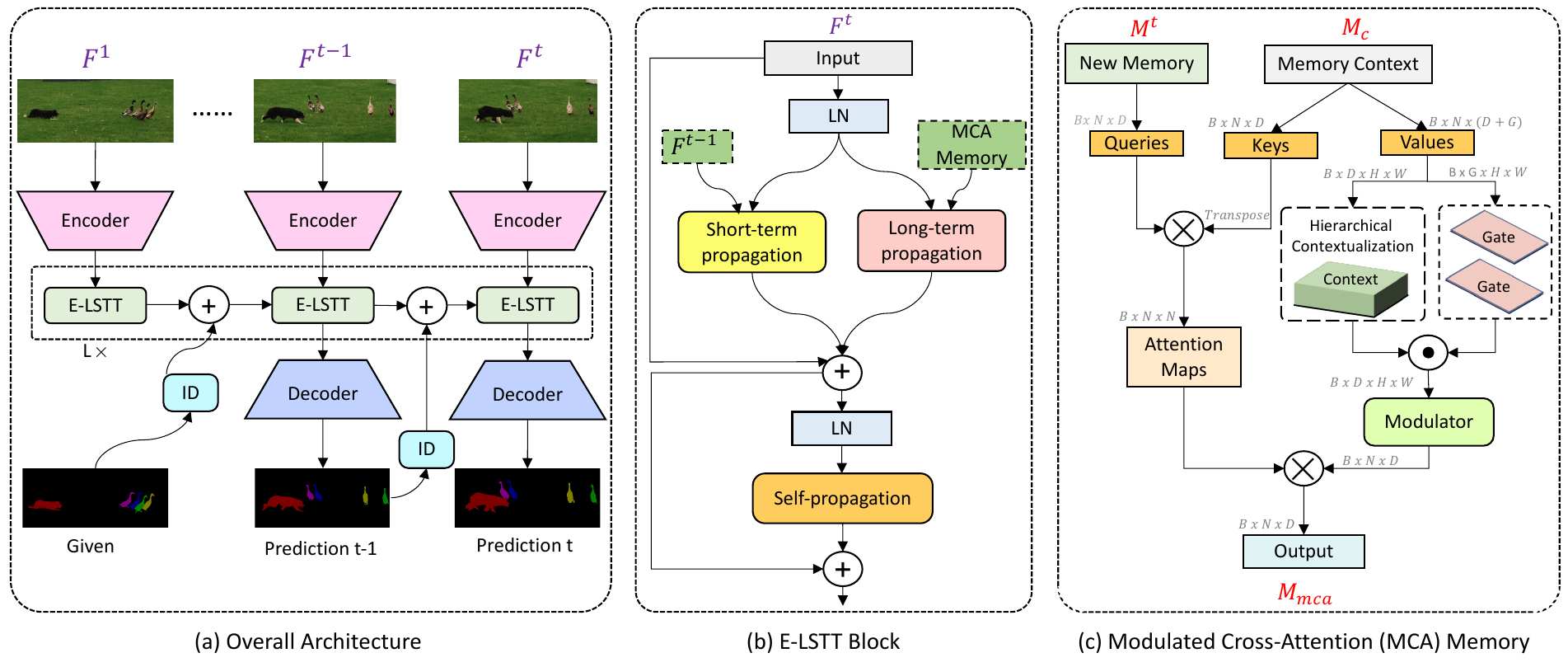}
    \caption{\textbf{Overview of the proposed MAVOS.} \textbf{(a)} An illustration for the overall architecture. The video frames are passed to the lightweight encoder to extract the frame features, followed by the proposed E-LSTT block to handle the long-term memory efficiently, followed by the decoder to generate the masks. \textbf{(b)} The details of the E-LSTT block. It mainly consists of short-term, long-term, and self-propagations to propagate the target information from the previous frames. The long-term propagation is based on the proposed Modulated Cross-Attention (MCA) memory. \textbf{(c)} Our proposed MCA memory. The new memory is projected to queries, and the memory context is projected to keys and values. We apply hierarchical contextualization using depth-wise convolution to generate the local context and multiply it by learnable gates. The aggregated context is projected and multiplied by the attention maps to generate the output.}
    \label{fig:method_overall}
\end{figure*}

\noindent\textbf{\textit{Optimized Memory}}: As discussed earlier, most existing transformer-based variants often struggle to segment long videos due to increasing demand for feature memory banks. Generally, these VOS models expand the memory with frame representation every few frames, rendering them impractical for videos with thousands of frames due to the heightened GPU memory usage. Conversely, other works~\cite{duke2021sstvos,mao2021joint} ignore long-term memory and focus only on short-term features. This omission limits the ability to encode long-term contextual information, hindering performance in challenging scenarios such as occlusion.
Unlike these approaches, an optimized long-term memory bank with consistent cost that can be flexibly integrated into a transformer-based VOS approach is desired. Such an optimized long-term memory bank is expected to capture the continuity of temporal contexts while maintaining consistent GPU memory consumption and speed, regardless of the number of frames.

\noindent\textbf{\textit{Effective Memory Encoding}}: 
In addition to high speed and consistent memory footprint, an effective long-term memory bank becomes crucial for precise mask prediction in long videos.  
To meet this requirement, the encoding mechanism of the memory should represent the features effectively and concisely. The goal is to establish meaningful and concise connections between memory frames at various time steps to effectively propagate the target information from past frames. This approach enables retention of relevant memory elements while progressively discarding irrelevant features from long-term memory over time.

\subsection{MAVOS Architecture}
 \noindent\textbf{Overall Architecture}: Fig.~\ref{fig:method_overall} (a) shows the overall architecture of our MAVOS, which is built on the recent transformer-based framework~\cite{yang2022deaot}. 
Given the video frames and the reference mask, the visual embeddings are extracted using the visual encoder. The identification assignment (ID) transfers the target masks into an identification embedding. Both visual and identification embeddings are propagated to the two branches of the proposed Efficient Long Short-Term Transformer (E-LSTT) block. The visual branch matches objects and gathers visual features from past frames. The ID branch reuses the attention maps of the visual branch to propagate the ID embedding from past frames to the current frame. Masks are predicted through the decoder with the same FPN~\cite{fpn}, as in~\cite{yang2022deaot}.  

\subsubsection{\textbf{E-LSTT Block}}
As shown in Fig.~\ref{fig:method_overall} (b), E-LSTT mainly consists of short-term propagation, long-term propagation, and self-propagation based on the gated propagation function of~\cite{yang2022deaot}. Short-term propagation is responsible for aggregating the spatial information of the target from the previous frame $\Fmat^{t-1}$. As discussed earlier, the baseline~\cite{yang2022deaot} employs long-term propagation to aggregate information about the target from all the stored previous frames; they append a new memory frame representation for each $\delta$ number of frames (set to 2/5 for training/testing). In contrast, here, the long-term propagation is re-designed to efficiently aggregate the information of the target from our MCA memory, which contains only the reference frame and single dynamic frame representation.
Assume the long-term memory contains a memory representation of the reference frame $\Mmat^{r} \in \R^{H \times W \times D}$, where $H$ and $W$ are the height and width of the feature map, respectively, and $D$ is the dimension of each token. After $\delta$ frames, a new memory representation $\Mmat^{t} \in \R^{H \times W \times D}$ added to the long-term memory bank, resulting [$\Mmat^{r}, \Mmat^{t}$]. At frame ${t+\delta}$, we update the long-term memory by replacing $\Mmat^{t}$ with $\Mmat^{t+\delta}$ based on following:
\begin{equation}
\Mmat^{t+\delta} = MCA(\Mmat^{t+\delta}, \Mmat^{t})    
\end{equation}
Finally, a self-propagation module is employed to learn the correlations among the targets within the current frame $\Fmat^{t}$.

\subsubsection{\textbf{Modulated Cross-Attention (MCA) Memory}}
\label{sec:method:cross_fusion:ours}
The objective of MCA memory is to propagate the information from the past frames to the target using a new fusion operator that can effectively handle both local and global features at various levels of granularity. 
With this objective, cross-attention~\cite{vaswani2017attention} and focal modulation~\cite{yang2022focal} can be potential choices. The cross-attention mechanism~\cite{vaswani2017attention} is a bidirectional interaction between two sequences; it allows the model to establish relationships and dependencies between the elements in two input sequences. Given the new memory frame $\Mmat^{t}$ and the memory context $\Mmat^{c}$, it uses a First Interaction Last Aggregation (FILA) process that initially involves producing queries from $\Mmat^{t}$ and keys and values from the context memory $\Mmat^{c}$. Attention scores are then calculated through the query and key interaction, followed by aggregation over the context. 
 Focal-Modulation~\cite{yang2022focal}, on the other hand, is a recent method for modeling token interactions that follows an early aggregation process by the first aggregation last interaction (FALI) mechanism.

Essentially, both self-attention and focal modulation involve the \emph{interaction} and \emph{aggregation} operations but differ in the sequence of function. In focal modulation, the context aggregation is performed first to $\Mmat^{c}$ through hierarchical contextualization and gated aggregation, followed by interaction $\mathcal{I}$ between the queries of $\Mmat^{t}$ and the aggregated features. The output of the aggregation is known as the \emph{modulator}, which encodes the surrounding context for queries. Note that the focal modulation is based on a hierarchy of convolutions that extracts localized features at various levels of granularity. However, this does limit the operator's capability to model global features as in cross-attention. 

Given the above-mentioned limitations of cross-attention and focal modulation, we propose a new fusion operator, named modulated cross-attention (MCA), that can handle local and global features at various levels of granularity. Consider the cross-attention formulation in Eq.~\ref{eq:joint:ca},
\begin{equation}
    \mathrm{CA}(\Mmat^{t}, \Mmat^{c}) = \mathrm{Softmax} \left( \frac{f_q(\Mmat^{t})f_k(\Mmat^{c})^\top}{\sqrt{d^k}} \right)f_v(\Mmat^{c}), \hspace{0mm}
    \label{eq:joint:ca}
\end{equation}
where $\mathrm{CA}$ indicates the cross-attention operator, $f_q$, $f_k$ and $f_v$ are the query, key, and value projections, respectively, and $1/\sqrt{d^k}$ is the scaling factor.

Similarly, we consider the focal modulation formulation, encompassing two steps: hierarchical contextualization, and gated aggregation, as shown in Eq.~\ref{eq:joint:fm}:
\begin{equation}
    \mathrm{FM}(\Mmat^{t}, \Mmat^{c}) = f_q(\Mmat^{t}) \circ f_{fm}(\mathrm{GA}(\mathrm{HC}(\Mmat^{c}))),
    \label{eq:joint:fm}
\end{equation}
where, $\mathrm{FM}$ is the focal modulation operator, $f_{fm}$ is the focal modulation projection, $\mathrm{HC}$ represents hierarchical contextualization and $\mathrm{GA}$ represents gated aggregation. In \emph{hierarchical contextualization}, the memory context $\Mmat^{c}$ is first projected by a linear layer $\Zmat^0 = f_{z}(\Mmat^{c}) \in \R^{H \times W \times D}$. Then, a series of depth-wise convolutions are applied to the projected feature map $\Zmat^0$ to produce $L$ refined feature maps, known as focal levels, as shown in Eq.~\ref{eq:hc}:
\begin{equation}
    \Zmat^{\ell} = \GeLU( \DWConv( \Zmat^{\ell-1} )) \in \R^{H \times W \times D},
\label{eq:hc}
\end{equation}
where $\Zmat^{\ell}$ is the feature map at focal level $\ell$, $\GeLU$ is the activation function, and $\DWConv$ is the depth-wise convolution operator. An additional global feature is obtained by average pooling the feature map $\ell = L$, given as $\Zmat^{L+1} = \AvgPool(\Zmat^L)$.
These feature maps are combined through a \emph{gated aggregation}. Gating weights are obtained by a linear projection of the memory context $\Mmat^{c}$ given by: $\Gmat = f_{g}(\Mmat^{c}) \in \R^{H \times W \times (L+1)}$, where $f_g$ is the gating projection. The gates $\Gmat$ are then multiplied with the feature maps at each focal level, followed by a summation to produce the aggregated feature map $\Zmat^{out}$ as given in Eq.~\ref{eq:joint:ga}:
\begin{equation}
    \Zmat^{out} = \sum_{\ell=1}^{L+1} \Gmat^\ell \circ \Zmat^\ell \in \R^{H \times W \times D},
\label{eq:joint:ga}
\end{equation}

Our proposed MCA mechanism utilizes a \emph{dual aggregation} design along with a cross-attention \emph{global interaction} style to effectively model both local and global features. 
Specifically, the MCA can be rewritten as Eq.~\ref{eq:joint:detailed}:
\begin{equation}
    \mathrm{MCA}(\Mmat^{t}, \Mmat^{c}) = \mathrm{Softmax} \left( \frac{f_q(\Mmat^{t})f_k(\Mmat^{c})^\top}{\sqrt{d^k}} \right)f_{fm}(\Zmat^{out}), \hspace{0mm}
    \label{eq:joint:detailed}
\end{equation}
where, $\mathrm{MCA}$ indicates the modulated cross-attention operator, $f_q$ and $f_k$ are the query and key projections used to calculate the attention matrix, $f_{fm}$ is the modulator projection, and $\Zmat^{out} = \mathrm{GA}(\mathrm{HC}(\Mmat^{c}))$ as given in Eq.~\ref{eq:joint:ga}. 

\begin{figure*}[t]
  \centering
    \includegraphics[width=0.9\linewidth]{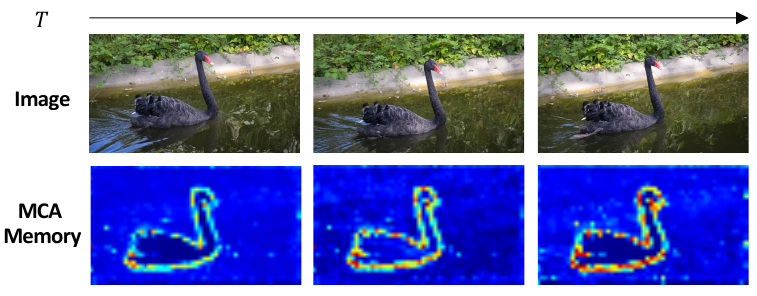}
    \caption{\textbf{Interpretation of the MCA memory}. Interpreting the dynamic frame of the MCA memory reveals its ability to encode temporal smoothness over time along the boundary of the black swan.}
    \label{fig:MCA_visualization}
\end{figure*}

The proposed MCA is an efficient memory mechanism with consistent memory usage and superior FPS, regardless of the number of frames. It enables the propagation of target information based on temporal changes in the past frames. The MCA memory encodes both local and global features at various levels of granularity. This helps establish effective connections between the memory frames and the target at different time steps, thereby facilitating target object segmentation in challenging situations. To interpret our MCA memory, we show in Fig.~\ref{fig:MCA_visualization} the visual representation of its dynamic frame. It illustrates that the MCA memory effectively encodes temporal smoothness over time.

\section{Experiments}
\subsection{Setup}

We evaluate our MAVOS and the most common VOS methods on three popular VOS benchmarks: DAVIS 2017~\cite{Pont-Tuset_arXiv_2017} for short videos, LVOS~\cite{voigtlaender2019feelvos}, and Long-Time Video~\cite{liang2020afb-ubr} for long videos. For a fair comparison, 480p resolution is used by default for all methods. We report the results using the three common evaluation metrics for VOS, Jaccard index $\mathcal{J}$ (similarity between predicted masks and ground truth), $\mathcal{F}$ score (average contour accuracy), and $\mathcal{J \& F}$ (the average of both values as the final score). 
We use the same pre-trained weights as DeAOT-L~\cite{yang2022deaot}, which is trained on the following static images~\cite{voc,coco,cheng2014global,shi2015hierarchical,semantic}. 
Then, we train our models on the same VOS benchmarks~\cite{Pont-Tuset_arXiv_2017, xu2018youtube} using 4 A100 GPUs. AdamW~\cite{adamw} optimizer is used for 100,000 steps with a batch size of 16, and a sequence length of 8 samples during training. The loss function is based equally on a combination of BCE loss~\cite{voigtlaender2019feelvos} and soft Jaccard loss~\cite{6909471}. The initial learning rate is set to 1 $\times 10^{-4}$ for DAVIS and LTV datasets, and 1 $\times 10^{-5}$ for LVOS. MAVOS updates the MCA memory per $\delta$ (set to 2/10 for training/testing) frames. Following ~\cite{yang2021aot,yang2022deaot,cheng2022xmem,voigtlaender2019feelvos}, we report the FPS using a single V100 GPU. The total training time is less than 25 hours.

\subsection{Datasets}

We briefly describe the utilized datasets, highlighting the total number of videos, mean frames per video, number of unique annotated objects, and total number of annotations across the entire dataset. The statistics are summarized in Table~\ref{tab:dataset_stats}.

\input{tables/dataset_stats}
\input{tables/LVOS}

\noindent{\textbf{Long-term Video Object Segmentation (LVOS)}}: This benchmark addresses the limitations of existing short-video benchmarks by introducing a more challenging long-video object segmentation dataset. Comprising 220 videos with 421 minutes, LVOS is the first densely annotated long-term VOS dataset. An average video duration of 1.59 minutes presents practical challenges, such as long-term object reappearing and cross-temporal similarities. Each video in LVOS consists of an average of 574 frames, encompassing 282 unique object categories and a total of 156,432 annotations. Furthermore, LVOS introduces highly challenging scenarios, including occlusion, absent objects, and different scales for objects, adding a layer of complexity to the evaluation. This dataset not only fills a crucial gap in long-term video object segmentation but also pushes algorithmic performance boundaries by simulating real-world conditions.

\noindent{\textbf{Long-Time Video (LTV)}}: It contains three videos with more than 7000 frames in total. Compared to the existing VOS benchmarks, this benchmark presents a significant contrast in scale per video. However, it lacks reliability with high variances of performances since it contains only three videos. Each video, with an average duration of 2470 frames, is evaluated based on 20 annotated frames sampled. This scarcity extends to the number of unique annotated objects, amounting to only three (one per video), with 60 annotations across the entire dataset. 

\noindent{\textbf{DAVIS 2017}}: It is one of the most used benchmarks in VOS. It is considered a short-term video object segmentation dataset since each video lasts a few seconds. The benchmark encompasses 90 training and validation videos, each comprising approximately 69 frames. With a focus on diverse challenges like occlusions and motion blur, DAVIS boasts 205 unique object categories and around 13,543 annotations in total. The significance of the dataset lies in providing high-quality, challenging, and multiple objects of interest. It is accompanied by densely annotated, pixel-accurate ground truth segmentation for each frame.

\subsection{Results on LVOS}

\begin{figure*}[t]
  \centering
    \includegraphics[width=0.9\linewidth]{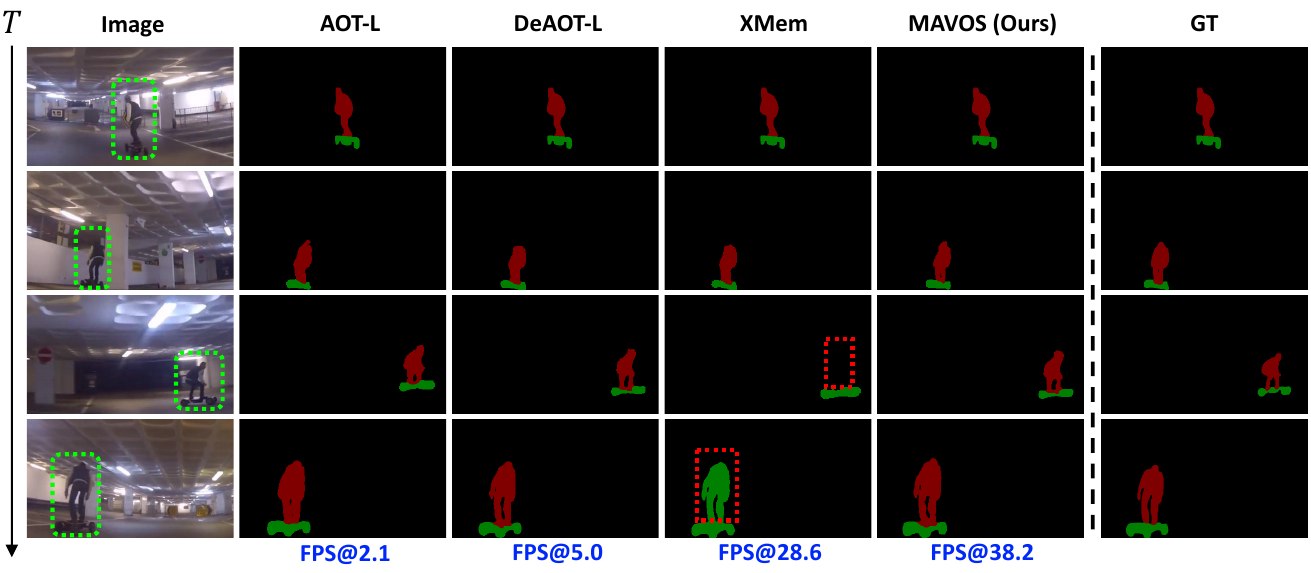}
    \caption{\textbf{Qualitative comparison between MAVOS and the SOTA methods on LVOS val set with the same backbone}. While the masks of AOT-L and DeAOT-L with infinite memory banks are promising, their real-time performance is hindered by a low FPS. On the other hand, XMem exhibits good FPS but struggles in challenging scenarios (marked in red dashed box). In the third row, XMem fails to recover the person occluded by the walls in the past few frames. In the fourth row, it confuses the person with the skateboard due to another disappearance. In contrast, our MAVOS accurately segments the targets despite the absence and occlusion with real-time FPS.}
    \label{fig:Qualitative_results}
\end{figure*}

\label{sec:results}

We compare our MAVOS with current state-of-the-art methods in Table~\ref{tab:LVOS} in terms of visual backbone, memory design, memory bank, $\mathcal{J \& F}$, and FPS. Our MAVOS achieves 60.9\% $\mathcal{J \& F}$ score, with an FPS of 38.2. In contrast, DeAOT-L achieves slightly higher $\mathcal{J \& F}$ at 61.1\%, with an FPS of only 5.0, and exploded GPU memory (see Fig.~\ref{fig:intro}). With ResNet-50, MAVOS achieves 63.3\% $\mathcal{J \& F}$ score at 37.1 FPS, which is significantly better than XMem by 13.3\% and 25\% faster. Compared to the state-of-the-art DDMemory, our MAVOS with the same backbone is 1.3$\times$ faster with better accuracy. Our SwinB variant of MAVOS outperforms all previous state-of-the-art methods by at least 3.7\% $\mathcal{J \& F}$. We show a qualitative comparison for AOT, DeAOT-L, XMem, and our MAVOS in Fig.~\ref{fig:Qualitative_results}.

\subsection{Results on LTV}
In addition to LVOS, we compare our MAVOS models with the current SOTA on LTV~\cite{liang2020afb-ubr} in Table~\ref{tab:LTV}.
Among existing works, XMem~\cite{cheng2022xmem} runs at 23.7 FPS with superior performance of 89.8\% $\mathcal{J \& F}$ score. Our MAVOS with SwinB achieves 90.3\% $\mathcal{J \& F}$ while running at 22.0 FPS. However, it is worth mentioning that LTV dataset consists of three videos only with 60 annotated frames.
Compared to the baseline transformers-based DeAOT-L, MAVOS achieves 87.4\% $\mathcal{J \& F}$ score, and reduces the GPU memory consumption from 38.6 GB to 4.9 GB. Also, the FPS increased from 4.1 to 38.9, without performance degradation.
\input{tables/LTV}
\subsection{Results on DAVIS 2017} To show that MAVOS also performs well on short videos, we further report results on the DAVIS 2017~\cite{Pont-Tuset_arXiv_2017} dataset in Table~\ref{tab:davis}. Our MAVOS achieves faster inference by nearly $10$ FPS compared to the baseline DeAOT-L~\cite{yang2022deaot}, with only a negligible $0.1$ drop in $\mathcal{J \& F}$ score. Similarly, our R50-MAVOS and SwinB-MAVOS achieves $85.6$\% and $86.4$\% $\mathcal{J \& F}$ score, respectively. This shows the generalization capability of MAVOS, which can model both short-term and long-term dependencies, allowing for strong performance on short and long-term video datasets at a significantly higher FPS and less GPU memory consumption.

\begin{figure}[t]
  \centering
    \includegraphics[width=0.92\linewidth]{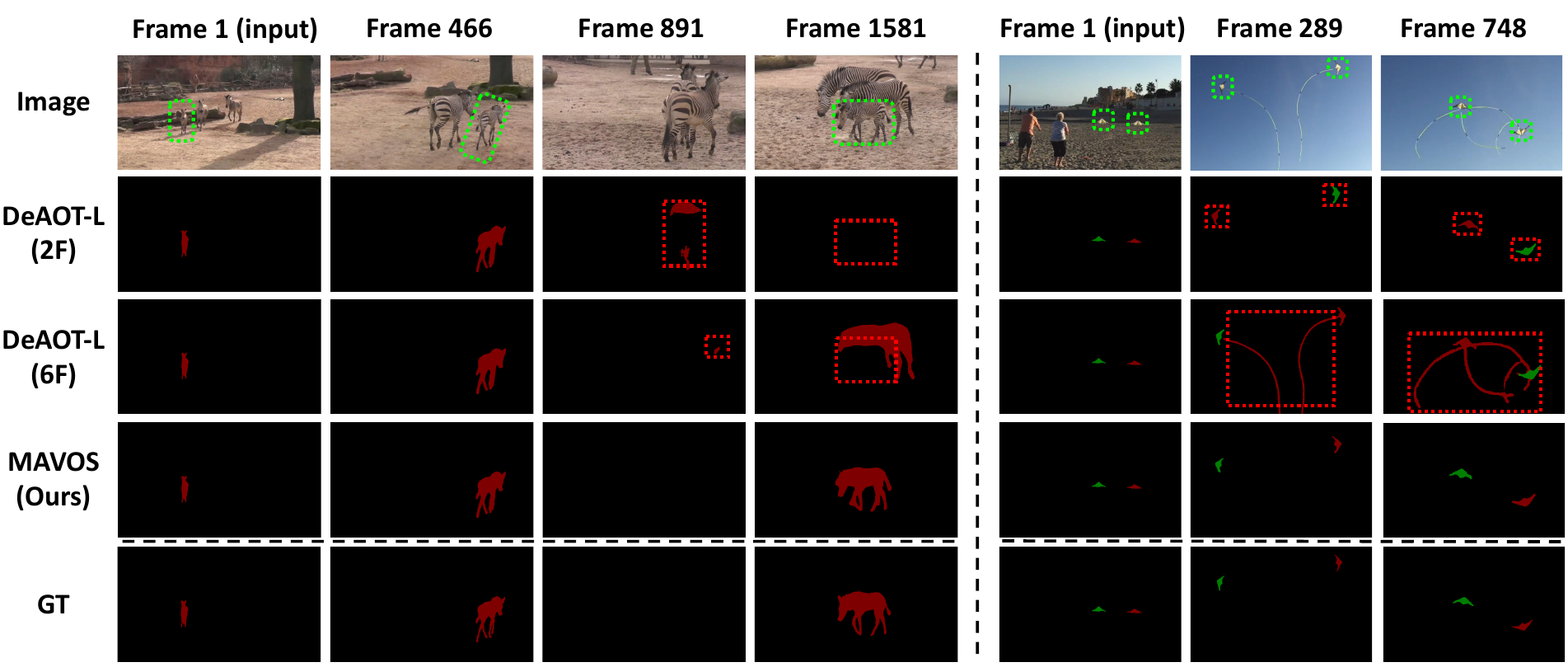}
    \caption{\textbf{Qualitative comparison between MAVOS and baseline DeAOT-L on LVOS val set}.
    \textbf{Left}: With two memory frames, the baseline struggles to correctly segment the target when it disappeared at frame 891 and reappears at frame 1581. Even with six memory frames, the baseline exhibits confusion between the target (small zebra) and potentially its mother (middle zebra) at frame 1581. \textbf{Right}: At frames 289 and 748, the baseline with two memory frames confuses both kites due to occlusion, and with six memory frames it over-segments the tails. In contrast, MAVOS shows impressive performance, accurately delineating targets despite the absence and occlusion.}
    \label{fig:Baseline_comparison}
\end{figure}

\subsection{Ablations} 
In this section, we ablate three sets of experiments to show the effectiveness of our MCA memory. In Table~\ref{tab:ablation:1}, we update the long-term memory bank of AOT-L and the baseline DeAOT-L to contain two frames (reference + previous) and six memory frames as proposed in~\cite{voigtlaender2019feelvos}. It is clear that with two memory frames, the performance of AOT-L and DeaOT-L is limited, specifically for long-term videos. With six memory frames, the performance of DeAOT-L is improved. However, our MAVOS, based on two memory frames, is even better than the baseline DeAOT-L with six memory frames on long-video datasets, with higher FPS. We also demonstrate that qualitatively in Fig.~\ref{fig:Baseline_comparison}.
\begin{table}[htbp]
    \footnotesize
    \centering
    \setlength{\tabcolsep}{3.0mm}
    \caption{Ablation with AOT-L and DeAOT-L with different number of memory frames on short and long-term benchmarks. FPS and memory are based on LVOS val set.}
    \begin{tabular}{lccc|cc}
        \toprule
        \multirow{3}{*}{Method} & \multicolumn{3}{c}{$\mathcal{J \& F}\uparrow$ }  & \multirow{3}{*}{FPS $\uparrow$} & \multirow{3}{*}{Mem (GB)$\downarrow$} \\
        \cmidrule(lr){2-4}
        & DAVIS  & LTV & LVOS  \\
        \midrule
        AOT-L-2F & 81.7 & 61.8 & 49.8 & 29.5 & 3.4 \\
        AOT-L-6F & 83.2 & 84.4 & 59.4 & 16.7 & 4.1\\
        DeAOT-L-2F & 82.5 & 81.2 & 57.2 & 39.1 & 3.3\\
        DeAOT-L-6F & \textbf{84.5} & 86.0 & 59.6 & 26.3 & 3.7 \\
        \midrule
        \rowcolor{LightCyan}
        \textbf{MAVOS (Ours)} & 84.4 & \textbf{87.4} & \textbf{60.9} & 38.2 & 3.3 \\
        \bottomrule
    \end{tabular}
    \label{tab:ablation:1}
\end{table}

\begin{table}[htbp]
    \footnotesize
    \setlength{\tabcolsep}{3.0mm}
    \centering
     \caption{Module-Level comparisons between our proposed MCA and other encoding methods on three short and long-term benchmarks.}
    \begin{tabular}{lccc}
        \toprule
        \multirow{2}{*}{Method} & \multicolumn{3}{c}{$\mathcal{J \& F}\uparrow$ } \\
        \cmidrule(lr){2-4}
        & DAVIS  & LTV  & LVOS \\
        \midrule
        Cross-Attention~\cite{vaswani2017attention} & 84.3 & 84.3 & 56.6 \\
         Focal-Modulation~\cite{yang2022focal} & 82.5 & 83.2 & 51.6 \\
         GCVIT~\cite{hatamizadeh2023global} & 83.4 & 84.9 & 55.1 \\
        \midrule
        \rowcolor{LightCyan}
        \textbf{MAVOS (Ours)} & \textbf{84.4} & \textbf{87.4} & \textbf{60.9} \\
        \bottomrule
    \end{tabular}
   
    \label{tab:ablation:2}
\end{table}
\begin{table}[htbp]
    \footnotesize
    \centering
    \setlength{\tabcolsep}{1.5mm}
    \caption{Ablation with different memory frames on short and long-term benchmarks.}
    \begin{tabular}{lccc|cc}
        \toprule
        \multirow{3}{*}{Memory Frames} & \multicolumn{3}{c}{$\mathcal{J \& F}\uparrow$ }  & \multirow{3}{*}{FPS $\uparrow$} & \multirow{3}{*}{Mem (GB)$\downarrow$} \\
        \cmidrule(lr){2-4}
        & DAVIS  & LTV & LVOS  \\
        \midrule
        Reference & 81.1 & 79.5 & 54.6 & 42.5 & 3.2 \\
        Previous & 80.2 & 74.6 & 37.3 & 42.9 & 3.2\\
        Reference + Previous & 82.5 & 81.2 & 57.2 & 39.1 & 3.3\\

        \midrule
        \rowcolor{LightCyan}
        \textbf{MCA (Reference + Dynamic)} & \textbf{84.4} & \textbf{87.4} & \textbf{60.9} & 38.2 & 3.3 \\
        \bottomrule
    \end{tabular}
    \label{tab:ablation:3}
\end{table}

In Table~\ref{tab:ablation:2}, for all experiments, we use the design of two memory frames (the reference frame and the dynamic memory frame) over the baseline DeAOT-L. First, when we replace MCA memory with cross-attention, it achieves comparable performance on DAVIS 2017 val. However, there is a significant decrease in performance on long-term video datasets LTV and LVOS by 3.1\% and 4.3\%, respectively. Secondly, using only focal modulation blocks with two levels instead of MCA memory results in decreased performance across all datasets, emphasizing the importance of the attention mechanism between target and memory frames. Lastly, we compare the MCA memory with GCVIT~\cite{hatamizadeh2023global}, a common 
technique for global-local feature encoding, and we show that MCA is better in the context of VOS. We also ablate in Table~\ref{tab:ablation:3} the effect of different long-term memory frames. The proposed MAVOS, based on MCA memory, achieves promising performance across all datasets compared to other memory frames.

\section{Conclusion}
\label{sec:conclusion}
Transformers have gained popularity in VOS applications due to their effective long-term modeling and propagation. However, with ever-increasing long-term memory, the GPU memory is exploding with very low FPS. In this work, we propose a novel mechanism to handle the long-term memory efficiently, while encoding useful temporal smoothness for the past memory frames. Our Modulated Cross-Attention (MCA) memory encodes only relevant information regarding the target and fades away the irrelevant information. Our MAVOS variant networks achieve favorable performance and generalize well on three VOS benchmarks (LVOS, LTV, and DAVIS 2017) with superior inference speed and less GPU memory consumption compared to other methods. Although MAVOS demonstrates effective segmentation in many scenarios, it may encounter challenges in accurately segmenting targets that are identical or highly similar, especially in cases involving sudden disappearance or severe occlusion. Addressing this specific challenge remains an area for future improvement.

\section{Acknowledgment}
\label{sec:Acknowledgment}
The computations were enabled by resources provided by the National Academic Infrastructure for Supercomputing in Sweden (NAISS) at Alvis partially funded by the Swedish Research Council through grant agreement no. 2022-06725, the LUMI supercomputer hosted by CSC (Finland) and the LUMI consortium, and by the Berzelius resource provided by the Knut and Alice Wallenberg Foundation at the National Supercomputer Centre.

\bibliographystyle{splncs04}
\bibliography{egbib}
\clearpage
\input{suppl_material}
\end{document}

%% file: mlVecMat.tex
\def\GeLU{\textsf{GeLU}} 
\def\DWConv{\textsf{DWConv}} 
\def\AvgPool{\textsf{Avg-Pool}} 

\newcommand{\beq}{\vspace{0mm}\begin{equation}}
\newcommand{\eeq}{\vspace{0mm}\end{equation}}
\newcommand{\beqs}{\vspace{0mm}\begin{eqnarray}}
\newcommand{\eeqs}{\vspace{0mm}\end{eqnarray}}
\newcommand{\barr}{\begin{array}}
\newcommand{\earr}{\end{array}}

\newcommand{\Fmat}[0]{{{\bf F}}\xspace}
\newcommand{\Gmat}{{\bf G}}

\newcommand{\Mmat}{{\bf M}}

\newcommand{\Zmat}{{\bf Z}}

\newcommand{\R}{\mathbb{R}}

\usepackage{color, colortbl}
\definecolor{Gray}{gray}{0.93}

\usepackage{xcolor,amsmath}
\usepackage[linesnumbered,ruled,vlined]{algorithm2e}
\DontPrintSemicolon

\SetKwComment{Comment}{\color{green!50!black}\# }{}

\SetKwProg{Function}{def}{:}{}

\SetKwProg{For}{for}{:}{}
\SetKwProg{If}{if}{:}{}

%% file: tables/dataset_stats.tex
\begin{table}[t]
    \footnotesize
    \setlength{\tabcolsep}{1.0mm}
    \centering
    \caption{
    Statistical comparison of the popular short video (DAVIS 2017~\cite{Pont-Tuset_arXiv_2017}) and long video (LTV~\cite{liang2020afb-ubr} and LVOS~\cite{voigtlaender2019feelvos}) VOS benchmarks. The largest value is in bold.}
    \begin{tabular}{l|c|c|c|c}
        \toprule
        Dataset & Videos  & Mean Frames per Video  & Objects & Annotations \\
        \midrule
        DAVIS~\cite{Pont-Tuset_arXiv_2017} & 90 & $\sim$69 & 205 & 13,543 \\
         LTV~\cite{liang2020afb-ubr} & 3 & \textbf{$\sim$2,470} & 3 & 60 \\
         LVOS~\cite{voigtlaender2019feelvos} & \textbf{220} & $\sim$574 & \textbf{282} & \textbf{156,432} \\
        \bottomrule
    \end{tabular}
    \label{tab:dataset_stats}
\end{table}

%% file: tables/LVOS.tex
\begin{table}[t]
    \setlength{\tabcolsep}{1.2mm}
    \centering
    \footnotesize
    \caption{Quantitative comparison with state-of-the-art models on LVOS validation set. MB denotes the kind of memory bank. OD, R+P, A, C, and MCA denote online adaption, reference and previous frames, all frames, compressed memory bank, and our modulated cross-attention memory, respectively. All methods are evaluated using a V100 GPU for a fair comparison. We report results in zero-shot settings on the validation set of LVOS. Our MAVOS performs favorably in terms of accuracy with superior speed against existing transformers-based methods.}
        \centering
        \footnotesize
            \resizebox{0.97\linewidth}{!}{%
        \begin{tabular}{l|c|c|c|ccc|c}
        \toprule
        Method& {Backbone} & {Design} & {MB}   & $\mathcal{J \& F} \uparrow$  &   $\mathcal{J} \uparrow$  & $\mathcal{F} \uparrow$  & {FPS $\uparrow$} \\ 
        \midrule
        LWL~\cite{bhat2020learning}  & ResNet-50~\cite{he2016deepResNet} & Template-based & OD    & 54.1  & 49.6        & 58.6  & 14.1            \\
        CFBI~\cite{yan2020cfbi} & ResNet-101~\cite{he2016deepResNet} & Template-based & R+P       & 50.0       & 45.0        & 55.1  & 5.2       \\
        \midrule
        STCN~\cite{cheng2021stcn} & ResNet-50~\cite{he2016deepResNet} & Memory-based & A    & 45.8        & 41.1        & 50.5  & 22.1            \\
        \multicolumn{1}{l|}{AFB-URR~\cite{liang2020afb-ubr}} & ResNet-50~\cite{he2016deepResNet} & Memory-based & C   & 34.8        & 31.3        & 38.2    &  4.8          \\
        RDE~\cite{li2022recurrent}  & ResNet-50~\cite{he2016deepResNet} & Memory-based & C    & 52.9        & 47.7        & 58.1   & 22.2            \\
        XMem~\cite{cheng2022xmem} & ResNet-50~\cite{he2016deepResNet} & Memory-based & C    & 50.0        & 45.5        & 54.4   & 28.6         \\ 
        
        DDMemory~\cite{Hong_2023_ICCV} & MobileNet-V2~\cite{sandler2018MobileNetV2} & Memory-based & C   & 60.7   & 55.0  & 66.3  & 30.3     \\
        \midrule
        AOT-B~\cite{yang2021aot} & MobileNet-V2~\cite{sandler2018MobileNetV2} & Transformer-based& R+P   & 53.4 & 47.7 & 52.1 &  26.6   \\

        AOT-L~\cite{yang2021aot}  & MobileNet-V2~\cite{sandler2018MobileNetV2} & Transformer-based& A   &  60.3  &     54.6    &    66.0  & 2.1         \\
        
        DeAOT-L~\cite{yang2022deaot} & MobileNet-V2~\cite{sandler2018MobileNetV2} & Transformer-based& A  & 61.1 &    55.2    & 67.1  &  5.0      \\
        
        \rowcolor{LightCyan}
        \textbf{MAVOS (Ours)}  & MobileNet-V2~\cite{sandler2018MobileNetV2} & Transformer-based& MCA &  60.9 &  54.6 & 67.1 & 38.2    \\
        \rowcolor{LightCyan}
        \textbf{MAVOS (Ours)} & ResNet-50~\cite{he2016deepResNet}& Transformer-based& MCA   & 63.3 &  57.6 & 69.0 & 37.1  \\
        \rowcolor{LightCyan}
        \textbf{MAVOS (Ours)} & Swin-Base~\cite{liu2021Swin} & Transformer-based& MCA  &  64.8    &  58.7    &  70.9  &  22.3   \\
        \midrule
        
        \end{tabular}
        }

\label{tab:LVOS}
\end{table}

%% file: tables/LTV.tex
\begin{table}[ht]
\setlength{\tabcolsep}{3.2mm}
\centering
\footnotesize
\caption{Quantitative comparison with state-of-the-art models on the three videos of the Long-Time Video benchmark. FPS is measured using V100 GPU.}
\begin{tabular}{l|ccc|c}
	\toprule
	Method & \mjf$\uparrow$ & \mj$\uparrow$ & \mf$\uparrow$ & FPS $\uparrow$ \\
	\midrule
        	CFBI+~\cite{yang2021cfbip} & 50.9 &  47.9 &  53.8  & 4.5 \\ 
	
	CFBI~\cite{yan2020cfbi} & 53.5 &  50.9 & 56.1 & 3.8 \\

	STM~\cite{oh2019videoSTM} & 80.6 & 79.9 &  81.3  & -\\ 
	MiVOS~\cite{cheng2021mivos} & 81.1  & 80.2 & 82.0 & - \\ 
	AFB-URR~\cite{liang2020afb-ubr} & 83.7 & 82.9 & 84.5 & 3.6 \\ 
	STCN~\cite{cheng2021stcn} & 87.3 & 85.4  & 89.2 & 20.1 \\ 
    XMem~\cite{cheng2022xmem} & 89.8  & 88.0  & 91.6  &  23.7 \\ 

    DeAOT-L~\cite{yang2022deaot} & 87.9 &  86.0 & 89.8 &  4.1 \\ 
    \midrule
    \rowcolor{LightCyan}
    \textbf{MAVOS (Ours)}& 87.4 &  85.7 & 89.0  & 38.9\\  
    \rowcolor{LightCyan}
    \textbf{R50-MAVOS (Ours)}& 87.5 &  86.1 & 88.9 & 38.2 \\ 
    \rowcolor{LightCyan}
    \textbf{SwinB-MAVOS (Ours)} & 90.3 &  87.7 & 92.9 &  22.0 \\ 

	\bottomrule
\end{tabular}

\label{tab:LTV}
\end{table}

\begin{table}[ht]
\setlength{\tabcolsep}{3.2mm}
\centering
\footnotesize
\caption{Quantitative comparison with state-of-the-art models on DAVIS 2017 validation set. FPS is measured using V100 GPU.}
\begin{tabular}{l|ccc|c}
	
    \toprule
     Method & $\mathcal{J\&F}\uparrow$ & $\mathcal{J}\uparrow$ & $\mathcal{F}\uparrow$ & FPS $\uparrow$ \\
    \midrule
    CFBI~\cite{yan2020cfbi} &  81.9  & 79.3  & 84.5 &  5.9  \\
    CFBI+~\cite{yang2021cfbip} &  82.9  & 80.1  & 85.7  & 7.2 \\
    STCN~\cite{cheng2021stcn} &  85.4  & 82.2  & 88.6  & 19.5  \\
    
    DDMemory~\cite{voigtlaender2019feelvos} &  84.2 & 81.3 & 87.1 & 28.1\\
    XMem~\cite{cheng2022xmem} &  86.2 & 82.9 & 89.5 & 24.2\\
    AOT-L~\cite{yang2021aot}  & {83.8} & {81.1} & {86.4}  & 20.3 \\
    DeAOT-L~\cite{yang2022deaot} & {84.5} & {81.6} & {87.4}   &  {29.3}  \\

    \midrule
    \rowcolor{LightCyan}
    \textbf{MAVOS (Ours)} & {84.4} & {81.5} & {87.2}  & {39.2}  \\
    \rowcolor{LightCyan}
    \textbf{R50-MAVOS (Ours)} & {85.6} & {82.6} & {88.5}  & {37.6}  \\
    \rowcolor{LightCyan}
    \textbf{SwinB-MAVOS (Ours)} & {86.4} & {83.2} & {89.6} & {21.8}   \\
    \bottomrule

\end{tabular}

\label{tab:davis}
\end{table}

%% file: suppl_material.tex
\clearpage
\setcounter{page}{1}
\setcounter{section}{0}

\begin{center}
  {\Large\textbf{Supplementary Material}}
\end{center}

\newcommand{\sectionbreak}{\clearpage}

\noindent{We provide additional details regarding}:

\begin{itemize}
    \item  Architecture Details of MAVOS 
    \item  Additional Ablation 
    \item  More Qualitative Results 
    \item  Limitations 
    \item  Discussion 
\end{itemize}

\section{Architecture Details of MAVOS}
\label{arch_impl} 
The baseline DeAOT~\cite{yang2022deaot} is designed with four network variants. DeAOT-T/S/B are tailored for short videos, they consider only the reference frame as the long-term memory, leading to consistent FPS and memory but poor accuracy on long videos because the temporal context is limited. DeAOT-L is designed for both short and long-term videos, it updates long-term memory by appending a new memory frame representation for each $\delta$ number of frames (set to 2/5 for training/testing). Since our motivation is to propose an efficient method for long-term videos, we introduce a single efficient network, called MAVOS, equivalent to DeAOT-L in terms of all hyper-parameters and the number of LSTT/E-LSTT blocks, which are set to three blocks.

We propose three variants of MAVOS based on three visual encoders (MobileNet V2~\cite{mobilenetv2}, ResNet-50~\cite{he2016deepResNet}, and Swin-Base~\cite{liu2021Swin}). Identification assignment (ID) of ~\cite{yang2022deaot} is used to transfer the target masks into an identification embedding. Both visual and identification embeddings are propagated to the two branches of the proposed Efficient Long Short-Term Transformer (E-LSTT) block. The visual branch matches objects and propagates visual features from previous frames. The ID branch reuses the attention maps of the visual branch to propagate the ID embedding from past frames to the current frame. The masks are predicted through the decoder using the same Feature Pyramid Network (FPN)~\cite{fpn}, as in~\cite{yang2022deaot}.

\section{Additional Ablation}
\label{additional_ablation}
In our evaluation on the LVOS validation set~\cite{voigtlaender2019feelvos} presented in Table~\ref{tab:ablation_focals}, we conduct an ablation to examine the impact of varying the number of focal stages in the proposed MCA memory. When employing two focal levels, R50-MAVOS achieves a performance of 63.3\% with a processing speed of 37.1 FPS. Notably, reducing the number of focal levels to one results in a marginal increase in FPS (1.4 FPS); however, this improvement is accompanied by a 0.8\% drop in performance. The introduction of a third focal level leads to a further decrease in FPS. This trend suggests that employing three focal levels may be less advantageous, potentially diverting attention towards high-level features at the expense of low-level features. This is less helpful here, since the MCA memory already encodes high-level features through the attention mechanism.

\begin{table}[h]
    \setlength{\tabcolsep}{3.0mm}
    \caption{
    Ablation for the number of focal levels in MCA memory of R50-MAVOS on LVOS validation set. The largest value is in bold.}
    \centering
    \begin{tabular}{c|cc}
        \toprule
        Focal Levels & $\mathcal{J \& F} \uparrow$ & FPS $\uparrow$  \\
        \midrule
        1 & 62.5\% & \textbf{38.5} \\
        2 & \textbf{63.3\%} & 37.1 \\
        3 & 63.1\% & 35.4 \\
        \bottomrule
    \end{tabular}
    
    \label{tab:ablation_focals}
\end{table}

\section{More Qualitative Results}
\label{qual_details}

\begin{figure}[t]
  \centering
    \includegraphics[width=0.97\linewidth]{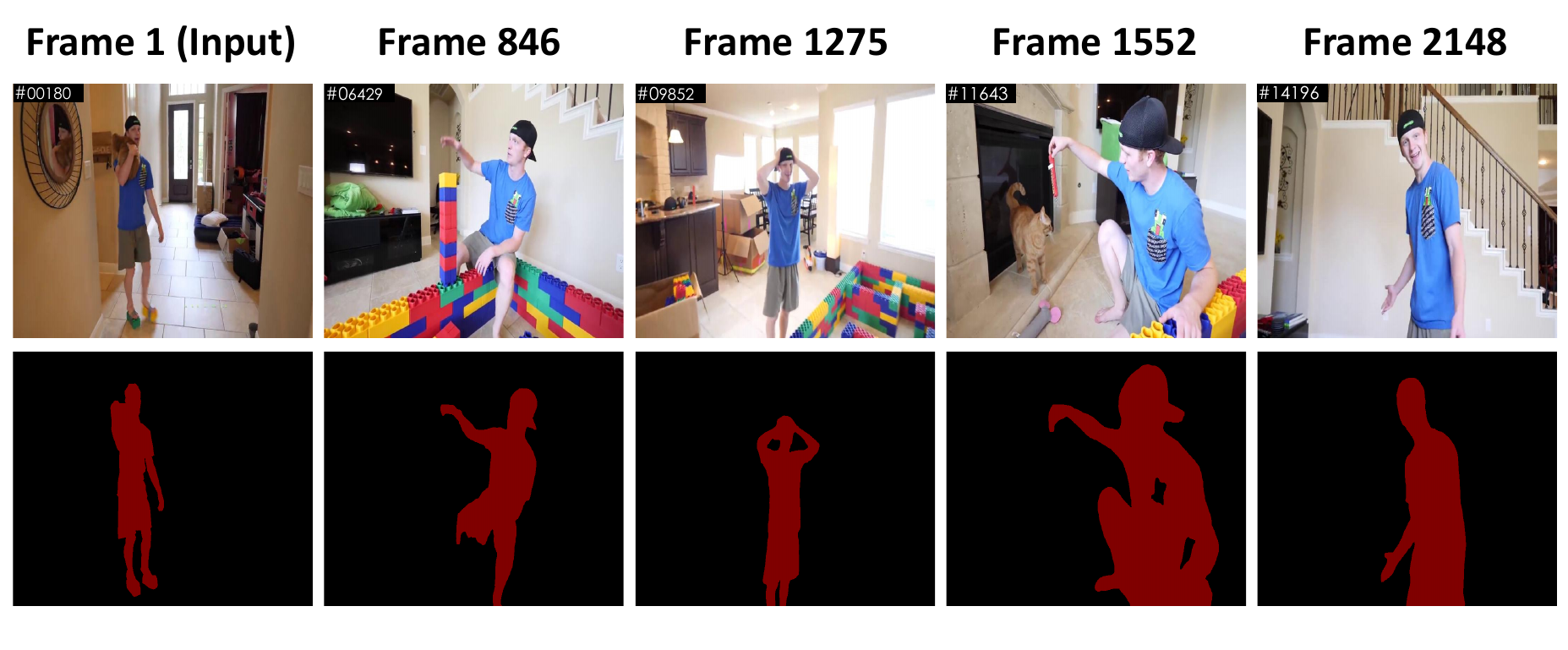}
    \caption{\textbf{Qualitative result for R50-MAVOS on the Long-Time Video dataset~\cite{liang2020afb-ubr}}. Our R50-MAVOS demonstrates good segmentation performance for sequences with more than two thousand frames at 38 FPS, accurately segmenting the target despite the fast movement.}
    \label{fig:suppl_qualitative}
\end{figure}
\begin{figure}[t]
  \centering
    \includegraphics[width=0.97\linewidth]{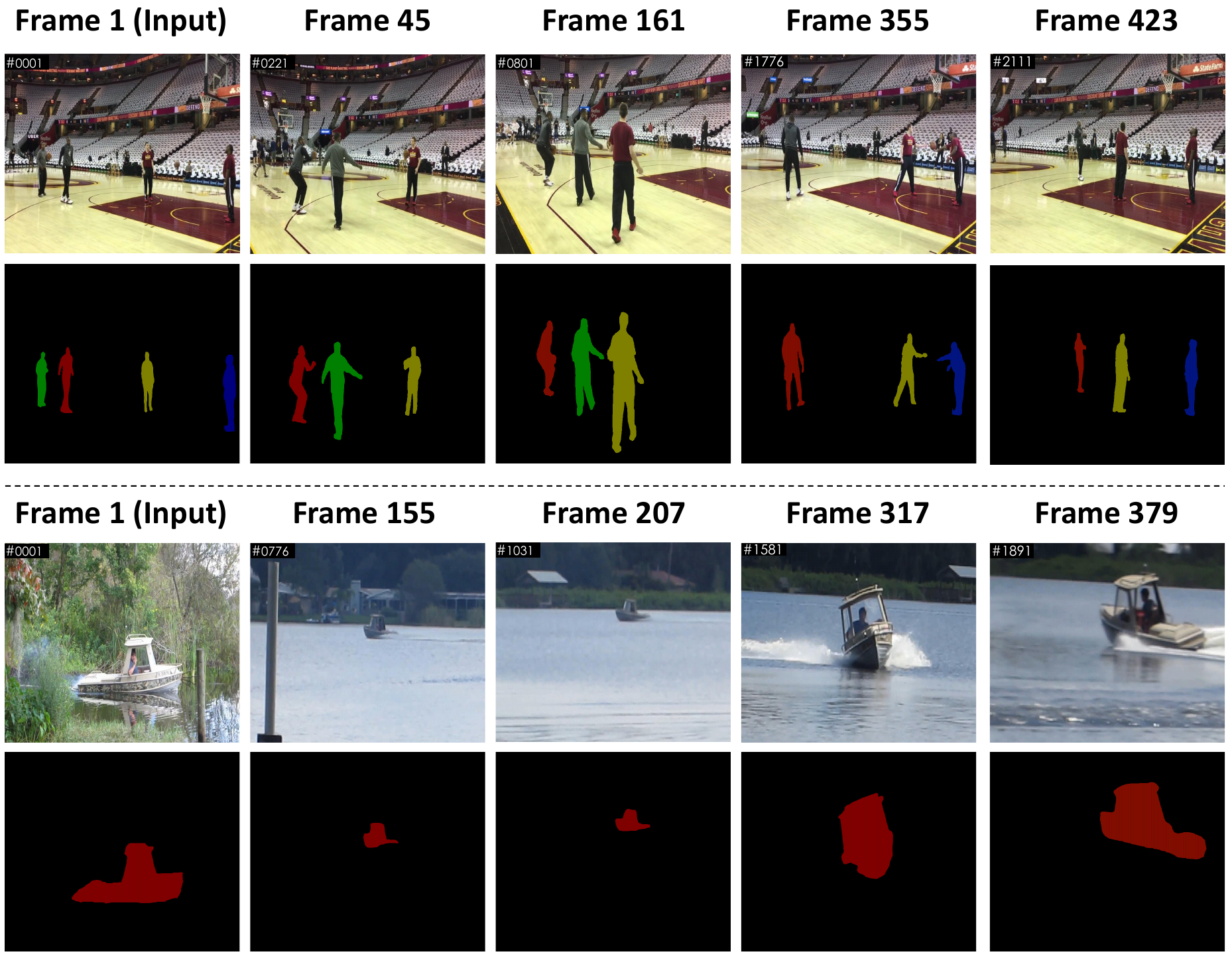}
    \caption{\textbf{Qualitative results for R50-MAVOS on two videos from LVOS validation set}. In the first two rows, the targets to segment are four basketball players in action. In the last two rows, the target is to segment the moving ship throughout the video. MAVOS showcases robust segmentation performance in both scenarios, accurately delineating targets despite occlusion and blocking in the first two rows, and coping with varying scaling factors in the last two rows.}
    \label{fig:suppl_qualitative_2}
\end{figure}

We show in Fig.~\ref{fig:suppl_qualitative} more qualitative results for R50-MAVOS on the Long-Time Video dataset~\cite{liang2020afb-ubr}. Our MAVOS demonstrates favorable segmentation performance for a long sequence (more than two thousand frames), and runs at 37 FPS. MAVOS accurately segments the target despite the fast movement of the boy and occlusion with other objects between the frames. In addition, we present in Fig.~\ref{fig:suppl_qualitative_2} more qualitative results on the LVOS dataset~\cite{voigtlaender2019feelvos}. In the first two rows, the given mask contains four different objects for segmentation throughout the video. Despite objects sometimes blocking each other and some disappearing and reappearing, Our MAVOS demonstrates promising performance for multi-object visual segmentation. In the last two rows, the goal is to segment a moving ship across the video. This is challenging because, firstly, the ship often moves far away, making it appear smaller. Secondly, the quality of this video is poor. Despite these challenges, our MAVOS is able to accurately segment the ship, whether it's close to the camera or far away, showing the effectiveness of our method.

\section{Limitations}
\label{limitations}
\begin{figure}[t!]
  \centering
    \includegraphics[width=0.97\linewidth]{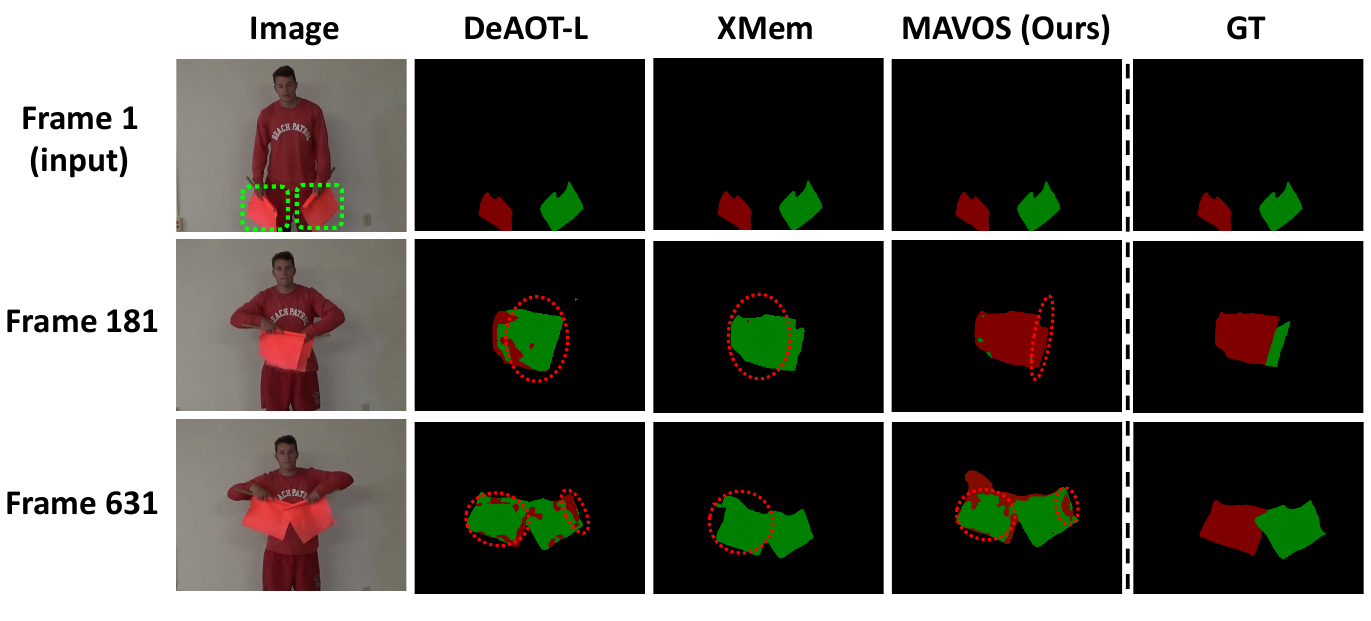}
    \caption{\textbf{Qualitative example for failure case from LVOS validation set}. MAVOS as well as the state-of-the-art methods fail to segment highly similar targets (almost identical) after severe occlusion. Both targets are marked with green dashed boxes, failure segmentations are marked with red dashed circles.}
    \label{fig:suppl_limitation}
    
\end{figure}

We observe that MAVOS often fails to segment targets when they are identical or highly similar after disappearance or severe occlusion occurs. We demonstrate this case in Fig.~\ref{fig:suppl_limitation}. This is a common problem not only for MAVOS but also for other state-of-the-art methods, including the baseline DeAOT-L~\cite{yang2022deaot} and XMem~\cite{cheng2022xmem}. This is likely due to the lack of encoding sufficient discriminative features for the targets due to the high similarity between them. As shown in Fig.~\ref{fig:suppl_limitation}, in the first column, two masks of almost identical flags are given in the reference frame. At frame 181, the flag with the red mask overrides the flag with the green mask. DeAOT and XMem confuse both flags, while MAVOS partially segments them correctly. At frame 631, MAVOS as well as DeAOT-L and XMem fail to discriminate both flags after the severe occlusion between both of them due to high similarity between both flags. We argue that this is likely due to the lack of discriminative features from the visual encoder and the short-term memory.

\section{Discussion}
\label{discussion}

Vision Transformers as end-to-end networks have gained popularity in video object segmentation applications due to their effective long-term modeling and propagation. However, their long-term memory is ever-increasing with exploded GPU memory and very low FPS. MAVOS runs at 38.2 FPS and achieves 60.9\% $\mathcal{J \& F}$ score. Our Modulated Cross-Attention memory is based on two memory frames only: the reference frame and one dynamic frame that encodes only relevant information regarding the target and fades away the irrelevant information. Our MAVOS variant networks achieve new state-of-the-art performance on three VOS benchmarks (LVOS, LTV, and DAVIS 2017) with superior run-time speed and significantly less GPU memory compared to the existing methods.